# Nerve Block Target Localization and Needle Guidance for Autonomous Robotic Ultrasound Guided Regional Anesthesia

Abhishek Tyagi, Abhay Tyagi, Manpreet Kaur, Richa Aggarwal, Kapil D. Soni, Jayanthi Sivaswamy and Anjan Trikha

*Abstract*— **Visual servoing for the development of autonomous robotic systems capable of administering UltraSound (US) guided regional anesthesia requires real-time segmentation of nerves, needle tip localization and needle trajectory extrapolation. First, we recruited 227 patients to build a large dataset of 41,000 anesthesiologist annotated images from US videos of brachial plexus nerves and developed models to localize nerves in the US images. Generalizability of the best suited model was tested on the datasets constructed from separate US scanners. Using these nerve segmentation predictions, we define automated anesthesia needle targets by fitting an ellipse to the nerve contours. Next, we developed an image analysis tool to guide the needle toward their targets. For the segmentation of the needle, a natural RGB pre-trained neural network was first fine-tuned on a large US dataset for domain transfer and then adapted for the needle using a small dataset. The segmented needle's trajectory angle is calculated using Radon transformation and the trajectory is extrapolated from the needle tip. The intersection of the extrapolated trajectory with the needle target guides the needle navigation for drug delivery. The needle trajectory's average error was within acceptable range of 5 mm as per experienced anesthesiologists. The entire dataset has been released publicly for further study by the research community at https://github.com/Regional-US/**

## I. Introduction

A peripheral nerve block (PNB) involves injecting a local anesthetic medication around a specific nerve(s) to block pain conduction. It is used to anesthetize a specific region of body such as an arm or a leg that needs surgery while avoiding general anesthesia and its associated complications [1].

Over the past decade, use of ultrasound (US) has dramatically increased the scope of PNB's and thus, regional anesthesia. Ultrasound guided regional anesthesia (UGRA) is a standard technique for administering PNBs as it offers anesthesiologists the ability to visualize the target anatomy and approach it in real-time. By allowing physicians to see the needle tip as it is being inserted, it helps ensure to maintain a correct trajectory and reduce the risk of any inadvertent trauma to other important structures such as blood vessel or other organs. US allows one to see the spread of medication as it is being injected. This ensures that the local anesthetic is distributed evenly, providing better pain relief while reducing overall local anesthetic medication volume. Thus, using US guidance for performing PNB's provides considerable advantages such as improved accuracy, safety, patient comfort and success rates and reductions in local anesthetic volume and complications [1].

The success of UGRA depends on accurate interpretation of sono-anatomy and maintaining an appropriate needle trajectory in real-time. This can often be daunting for novice trainees and serves as an impediment to wide scale adoption of UGRA [2]. Moreover, the global shortage crisis and widespread inequity of health care workers makes the development of autonomous robotic systems capable of administering UGRA desirable for remote settings where anesthesiologists are unavailable [3]. Real-time segmentation of nerves, needle tip localization and needle trajectory extrapolation are some of the requirements for developing such a system. Nerve segmentation is required for defining the target anatomy. The nerves in the US videos are not stationary in nature inherently due to movements of US probe during scanning. The insertion of needle and its movement through deformable tissues applies pressure on adjacent anatomical structures and changes their relative positions, necessitating a visual servoing based control system. The real-time needle tip localization and needle trajectory are important components for needle guidance and path-planning.

Ultrasonography in general is considered a challenging modality for image segmentation because of poor contrast images, operator dependency, and the reduced SNR by granular image texture from speckle noise. In addition, for real-time segmentation, the hazy boundaries and speckle noise introduce additional time-complexity hurdles on image preprocessing. Needles occupy very few pixels in images which pose additional challenges as their appearance is often degraded by reverberation, comet tail, side-lobe, beamwidth, or bayonet artifacts [2]. When the needle is not aligned parallel to the plane of the ultrasound beam, it will only be partially visible where it intersects with that plane. Additionally, the needle may be mistaken for nearby linear anatomical structures, such as the edges of bones or pleura. The invasive nature of needles limits the possibility of acquiring large scale data required for deep learning methods.

For humans, among the regional anesthesia targets for upper limb surgeries, two of the most commonly employed techniques include brachial plexus blockade at Interscalene

Abhishek Tyagi was with International Institute of Information Technology, Hyderabad, India and is now with Asian Institute of Gastroenterology, Hyderabad, India (corresponding author: mr[dot]tyagi[at]gmail.com)

Abhay Tyagi MBBS, M.D. is with Brigham & Women's Hospital, Harvard Medical School, Boston, MA, USA

M. Kaur and A. Trikha were with Dept. of Anesthesiology, All India Institute of Medical Sciences, New Delhi, India. They are now with Dept. of Anesthesiology and Perioperative Medicine, Penn State Milton S Hershey Medical Center, Hershey, Pennsylvania, USA.

R. Aggarwal and K. D. Soni are with Dept. of Anesthesiology, All India Institute of Medical Sciences, New Delhi, India

J. Sivaswamy is with International Institute of Information Technology, Hyderabad, India

space (ISC) and supraclavicular space (SCBP) [4]. The brachial plexus is a network of nerves that branches off the spinal cord in the neck and supplies the arm, shoulder, and hand. The ISC block is performed at the level of nerve roots exiting from the cervical spinal cord and provides effective anesthesia to shoulder region and upper arm whereas SCBP block is performed below, lateral to the clavicular head of the sternocledomastoid muscle and provides analgesia to the upper arm, forearm and hand [4]. When performing these blocks under real-time US guidance, the clinician looks for distinct sono-anatomical patterns, such as, cluster of hypoechoic nodules or bunch of grapes next to subclavian artery for SCBP and a vertical chain of hypoechoic nodules or a traffic light sign sandwiched between two scaleni muscles for ISC [9]. However, the aforementioned patterns only serve as general guide and can vary considerably depending upon individual patient's anatomy and US probe orientation. Moreover, these two sites are situated very close to each other (within a few centimeters) in the neck and transition seamlessly into one another making it challenging to distinguish for novice operators. SCBP blockade, even though closely anatomically related to ISC brachial plexus, provides anesthetic coverage to a distinct area and is devoid of important side effects related to ISC block such as phrenic nerve paralysis and Horner syndrome [4].

This research focuses on the computer vision related aspects for developing an autonomous UGRA system in two stages. First we created a large high fidelity dataset and developed a deep learning model capable of segmenting SCBP and differentiating it from ISC in neck US videos. Then, we defined automated anesthesia needle targets derived from CNN nerve segmentation predictions and designed and implemented automated needle segmentation, needle tip localization and needle trajectory extrapolation to enable a visual servoing based control system.

As part of the outcomes of this research, we also publicly released a dataset of 227 brachial plexus US videos (41,000 frames) with segmentation annotations for all frames from experienced anesthesiologists and needle annotations of 6 SCBP videos comprising of 1,061 frames. To the best of our knowledge, this is the largest publicly released dataset of annotated needle frames in US videos created from human subjects. This is also the first study on simultaneous identification of the target anatomy as well as targeting needle in given nerve US videos.

## II. Related Work

Hemmerling et al [6] demonstrated a tele-operated robotic PNB on the sciatic nerve in a 2013 pilot study, however, the nerve was identified manually. In recent years, attempts have been made to identify individual nerves, surrounding structures and needles using object detection and segmentation neural networks [7, 8]. Utilizing the 5,365 ISC training images of Kaggle dataset [9], Baby et al. [10] and Kakade et al. [11] implemented the UNet model to achieve dice coefficients of 0.71 and 0.69 respectively, but evaluation on the test dataset could not be done since the test dataset annotations were not released by Kaggle. Both of these approaches utilized time consuming image-processing steps that made the real-time inference infeasible. Wu et al improved upon this with Region-aware pyramid aggregation model and Adaptive pyramid to achieve a dice coefficient of 0.74 at real-time performance [12]. Ding et al. [13] created a new dataset of 1,052 frames and publicly released its training dataset of 955 frames with annotations for ISC and surrounding anatomical structures. They improved upon Mask-RCNN utilizing the spatial information of surrounding anatomical structures and achieved a dice score of 0.51 at 5.19 FPS. Smistad et al. [14] used U-Net [15] to delineate individual brachial plexus nerves in axillary space and reported an F-score ranging from 0.39 (radial nerve) to 0.73 (median nerve) using 25% IoU overlap criterion with real-time inference capability. Most of these studies are limited by time complexity and small-sized training datasets that affects their accuracy and generalizability.

For needle localization, the literature review [7, 16] shows early attempts with classic image processing methods such as hough transform, thresholding, top-hat filter, Gabor filter, Kalman filter and Radon transformation. In the last decade, researchers improved upon them with machine learning based methods. Most of the techniques for automatically detecting needles often rely on phantoms or ex-vivo tissues from chickens and pigs [7]. However, these materials fail to capture the complexity of human tissues, which are more varied and possess unique features absent in models. This difference highlights the essential need for large, diverse datasets from real human data to enhance the accuracy and effectiveness of needle detection techniques.

The studies with human data are limited in number due to the difficulty in acquiring data because of the invasive nature of needles. Hatt et al. [17] worked on 577 images from 6 patients assuming known needle orientation and reported a trajectory distance error of 0.2 mm with a processing time of 3 seconds using Log-Gabor wavelets. Lee et al [18] worked on 996 images from 8 patients and achieved a trajectory angle error of 13$^{o}$ and trajectory distance error of 9 pixels at 3 fps using an excitation network.

## III. Materials and Methods

### A. *Dataset Curation and Annotation*

This was a prospective study conducted jointly at the All India Institute of Medical Sciences, New Delhi, India and the International Institute of Information Technology, Hyderabad, India. Institutional ethics committee approval was obtained prior to commencement of the study (IEC/18/1/2020,RP-02/2020). Patients of age 20-80 years were randomly selected to be included in this study. Exclusion criteria were patients known to have brachial plexus injury, brachial plexopathy, cervical trauma, limitations to neck movement or patient refusal. After obtaining written informed consent from patients, they were scanned with a linear high frequency probe of 12-15 MHz of US scanner eSaote MyLab One™ (eSaote SPA, Genoa, Italy). Subjects were scanned on a 30-45$^{O}$ incline, with the head turned away from the side of interest with the overarching goal of focusing on the brachial plexus in the SCBP and ISC space. US probes were then tilted, rotated and slid around in the cranio-caudal and medial to lateral directions to capture dynamic brachial plexus patterns in addition to adjoining areas of the neck such as muscles, fascia and blood vessels.

Random low, medium and high gain as well as different depth settings were used to increase the variation in the dataset. Table I summarizes the patient demographic characteristics. All personal identifiable information was cropped. The resulting videos were 542x562 pixel resolution.

Annotation

A custom video annotation tool was built using OpenCV software and is available on the GitHub. In a given video, a random seed frame was annotated for SCBP by an expert anesthesiologist using bounding boxes. The bounding boxes were drawn to cover the approximate SCBP area and were not meant to annotate individual nerve roots. The boxes were drawn to stay clear of critical structures such as the subclavian artery and pleura. Subsequent frames were automatically annotated using the CSR-DCF and KCF object-tracking algorithms [19,20]. Each automated annotation was either approved or rejected by the annotator. If rejected, the bounding boxes were redrawn as new seeds for automated annotations. The individual boxes were then fused to create a single mask. The contour of this mask was then shrunk to better match the boundaries of the plexus using morphological geodesic active contour algorithm. The parameters of the algorithm were chosen for each individual video separately and are available in the dataset. The final shrunken masks were labeled as "ground truth" masks (Fig.1) which were used to train the neural network. The frames in which the primary annotator was uncertain, a second expert was called in to review the frames. If still uncertain of the presence or absence of SCBP or ISC, these frames were discarded to avoid noisy annotations.

For the needle segmentation, six videos consisting of 1,061 frames containing both the needles and SCBP were used. The coordinates of the needle tip and needle tail-end at the skin were annotated by experienced anesthesiologists. Four pixel thick white lines between these coordinates were drawn to create needle masks.

Data Augmentation

In addition to the standard data augmentations methodologies of horizontal flip, rotation and gamma intensity transformation; synthetic acoustic shadows were also used. US can create acoustic shadows when air or tissue inhibit sound waves from penetrating deeper into the body. A synthetic shadow is generated using a 2D Gaussian where the standard deviation determines radius of the shadow and the kernel coefficient determines the intensity of the shadow. Augmented image is obtained by multiplication of Gaussian to the image.

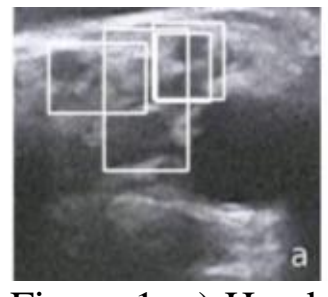

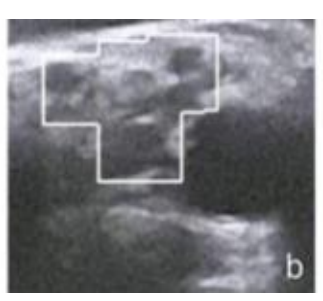

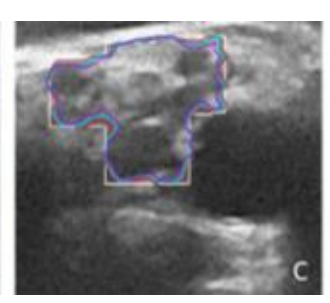

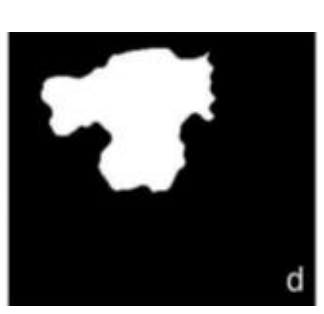

Figure 1. a) Hand drawn annotation bounding boxes to cover the plexus as a whole while avoiding critical structures such as the pleura and subclavian artery b) outer boundary of the union of boxes selected as the initial contour c) Various shrunken contours proposed by the morphological geodesic active contour algorithm using different parameters d) Best contour approximating the boundaries of the plexus chosen by experienced anaesthesiologists as the final ground truth

TABLE I. DATASET PATIENT CHARACTERSTICS

| Characteristic | Total | Supraclavicular | Interscalene |
|---|---|---|---|
| No. of Patients | 196 | 157 | 33 |
| *Male* | 134(68%) | 107(68%) | 23(70%) |
| *Female* | 62(32%) | 50(32%) | 10(30%) |
| No. of Videos[a] | 196 | 157 | 33 |
| Total Frames | 34,926 | 29,129 | 4,622 |
| *Positive*[b] | 26,966 | 23,302 | 3,664 |
| Avg. Age (years) | 42(13.6) | 42(13.5) | 42(13.9) |
| Avg. Height (cm) | 170(6.1) | 171(5.7) | 169(6.8) |
| Avg. BMI | 25.3(2.6) | 25.4(2.6) | 25.2(2.7) |

Data is reported as n (%) or mean (±SD)

a. 6 videos contained only negative data and used only for training.

b. Positive data frames imply the presence of SCBP or ISC in a given frame.

## *B. Nerve Segmentation Neural Network*

From a wide range of available off-the-shelf segmentation models, we shortlisted two models: DeepLabv3 [21] and UNet [15], and their enhancements: DeepLabv3+[22] and UNet++ [23] based on established usage in medical image segmentation literature [7, 8, 16, 24], real-time processing capability and low memory usage. All the four models use 50-layer residual-net as backbone.

One of the limitations while using traditional fully CNN for image segmentation tasks is that the spatial resolution is progressively downsampled resulting in information loss and low resolution output masks with fuzzy object boundaries. The DeepLabV3 model addresses this by using Atrous convolutions and Atrous Spatial Pyramid Pooling which help in extracting information with a wider receptive field and thus avoiding the loss of spatial resolution. The U-Net model mitigates the downsampled resolution problem through up-convolutions on the expanding path and concatenations with features from the downsampling path, resulting in a symmetrical U shaped network.

For internal model validation, stratified 5-fold cross-validation was performed. The training:validation:test proportions were 61:19:20 for each fold. The stratified samples were chosen to keep roughly equal proportions for dexterity (left/right), frequency gains (high/medium/low), and gender (male/female) in each of the training, validation and test sets.

## *C. Nerve Segmentation Model Generalizability*

For testing generalizability of the model, two separate datasets of 15 videos each were constructed using different US machines by different operators. The first dataset was created using Sonosite M-Turbo™ (Fujifilm Sonosite Inc, Bothell, WA, USA) which produces videos of similar quality to the original machine in the main dataset. Six of the 15 videos were acquired with needle in-situ to test the performance of the model in the presence of needle. These 15 videos contained a total of 2,186 frames. The second dataset was created using a relatively inexpensive portable handheld machine Butterfly-IQ™ (Butterfly Network Inc., USA) which produces videos of lower quality. The 15 videos from the second machine contained a total of 3,088 frames. Models were evaluated on these datasets with and without fine-tuning.

*D. Nerve Block Target*

The SCBP lies adjacent to the subclavian artery which can be identified due to its pulsating nature. Underneath the plexus, lies the clavicle bone and pleura or the outer layer covering the lungs. A successful block involves charting the needle course to deposit the drug below and above the plexus as well as preventing any injury to the adjacent artery and the pleura.

In order to define the needle target, first an ellipse is fitted on the contour of the nerve segmentation prediction. The top and bottom peripheries of the nerve plexus are defined as elliptic arcs between +/- 30 degrees from a vertical line through the center of the ellipse regardless of the orientation of the ellipse. Finally, the anesthesia target is defined as the region 4 pixels (0.8 mm) above and below these elliptic arcs as shown in Fig. 2.

*E. Needle Segmentation and Trajectory*

The strategy was to repurpose the neural network trained on the US domain (Section III-B) wherein the network learned to segment SCBP and ISC as two output channels. In this step, a second fine-tuning on five needle videos was done to adapt the second output channel of ISC to predict needle mask. After training on five random videos we tested the needle in the sixth video.

While real-time ultrasound can accurately locate areas within the body where needles are present, it may struggle to visualize the entire needle, especially when inserted at a steep angle. To address the problem of partial needle visibility, the visible portion was used to extrapolate the trajectory. Radon transform was applied on the network's needle mask prediction to obtain a sinogram. This transform essentially comprises projections at different angles.

The data obtained from the Radon transform is commonly referred to as a sinogram, as the Radon transform of a point source that is not in the center, produces a sinusoidal shape. Assuming that the sinogram's maximum value corresponds to the needle trajectory's angle and that the lowest point in the needle mask is the needle tip, a line was drawn from the needle tip to show the needle trajectory.

As depicted in Fig. 3 and supplementary video, when the needle trajectory intersects with the target anesthesia region,

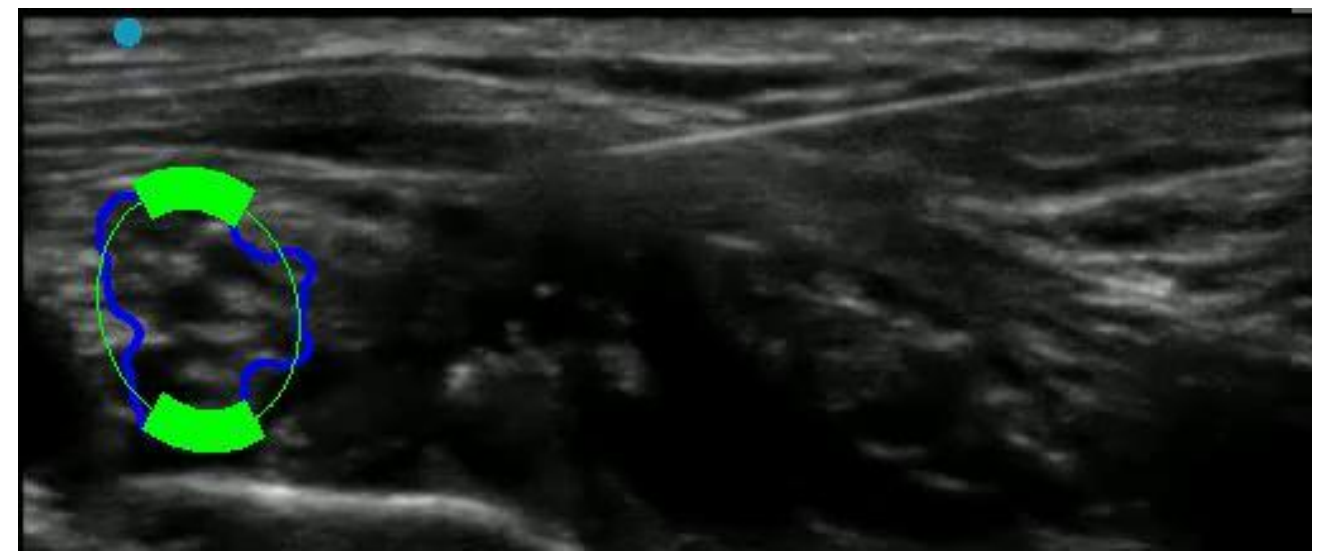

Figure 2. An ellipse (green) is fitted on the contour (blue) of the nerve segmentation prediction. Needle target (green thick arcs)

it is colored green. When it is between the anesthesia targets, it is colored yellow and is colored red if it misses the anesthesia targets.

*F. Evaluation*

As the hazy boundaries of the nerve plexus are not perfectly delineated in US videos, the goal of this study was not pixel-accurate segmentation but rather to identify the approximate location of the plexus in the US images. Therefore, the performance evaluation was done using both object detection metrics and segmentation metrics.

For object detection metrics, the precision, recall and F-1 scores were calculated using intersection over union (IoU) thresholds. The predictions too small to be regarded as SCBP or ISC were discarded. The threshold for this was set at 20% of the median area occupied by the ground truth in the entire dataset, i.e 3240 pixels for median SCBP area and 914 pixels for ISC in resized 256x256 images.

The different spatial resolutions of different US machines were brought to common ground of width 256 pixels for comparison. The original dataset and Butterfly-IQ™ dataset were evaluated at 256x256 resolution and Sonosite M-Turbo™ dataset were evaluated at 256x192 resolution.

For segmentation metrics, dice coefficient was calculated in three different ways:

- Dice for both plexuses in all frames of all videos: This measured the ability of the model to differentiate the two plexuses. When the model correctly identified the absence of nerve in any image i.e. in the true-negative case, 100% dice-score was assigned for that image. However, this inflated the dice coefficient.
- Dice for SCBP was calculated using only SCBP videos and dice coefficient for ISC was calculated using only ISC videos. Here also, the frames with true-negative cases

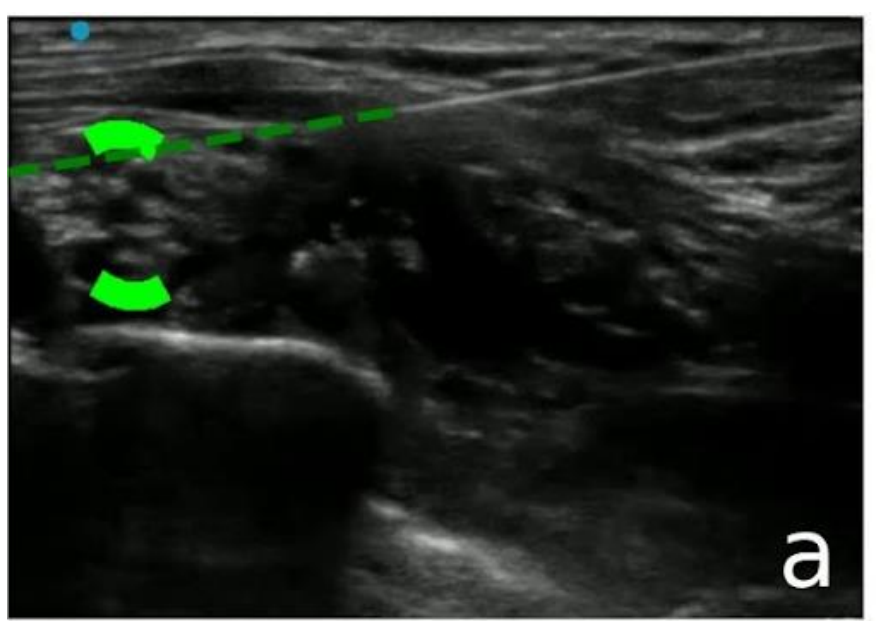

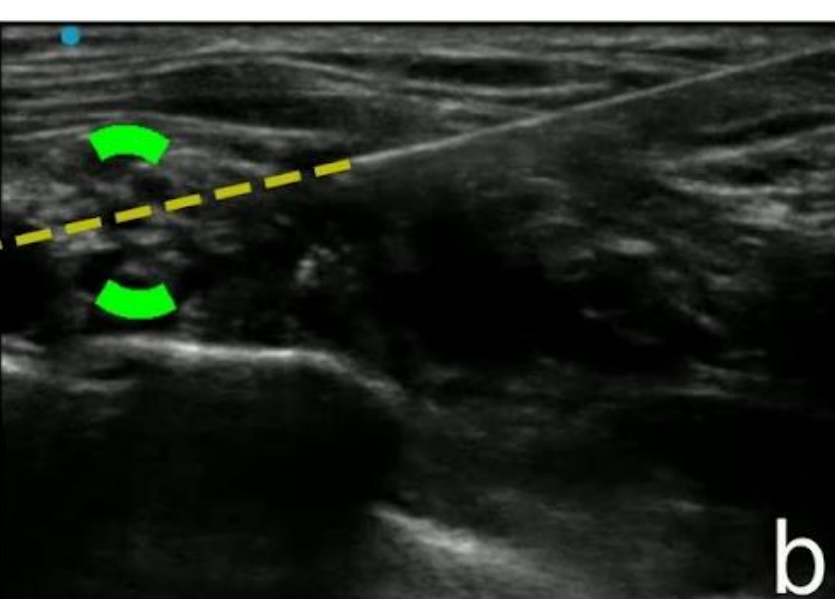

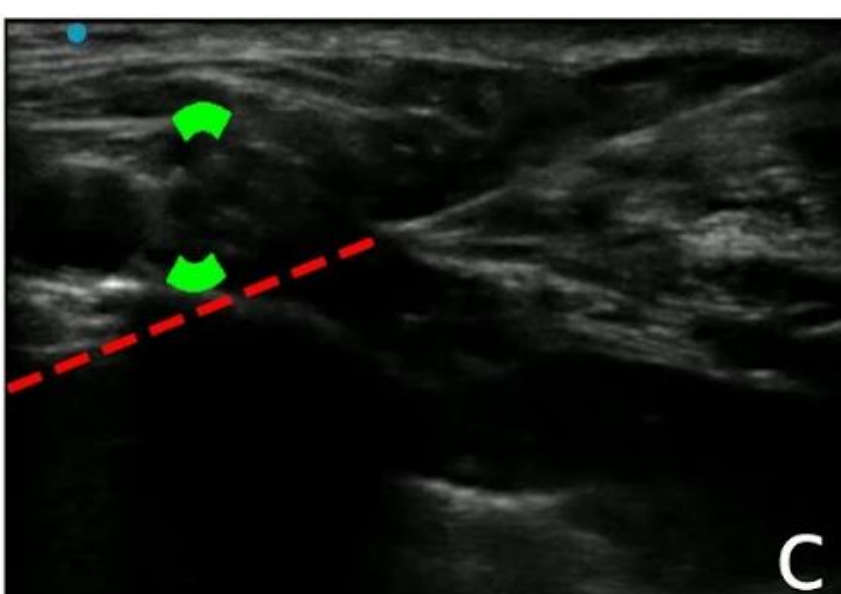

Figure 3. Needle Trajectory guidance towards regional anesthesia target.

inflated the overall dice scores for the video.

- Dice calculated separately for SCBP and ISC using only the positive frames in SCBP and ISC videos respectively. The frames with true-negative cases were ignored and not taken for consideration.

For all the above metrics, the average metric of all considered videos is the final metric. The metric for individual video is the average metric of all frames in that video.

The needle trajectory was evaluated by comparing the ground truth and prediction for trajectory angle, trajectory's distance from center of the image, and needle tip coordinates and average errors over all frames in a video were reported.

## IV. Results

### A. Nerve Detection

For a total of 163 videos, six did not contain SCBP and were used for training purposes only. In total, 29,129 image frames were annotated with 23,302 positive frames (SCBP present) and 5,827 negative frames (SCBP absent). Table II, shows the average precision, recall and F score ± standard deviation and three dice coefficients as described in the Evaluation section for the four baseline models' performance on SCBP and ISC. Table III shows the performance of UNet++ model before and after fine-tuning on the datasets created through two other US scanners.

TABLE II. Object Detection & Segmentation Models

| | **DeepLabV3** | **DeepLab V3+** | **UNet** | **UNet++** |
|---|---|---|---|---|
| ***Supraclavicular BP (SCBP)*** | | | | |
| Object Detection at 50% IOU | | | | |
| Precision | 0.86(0.22) | 0.87(0.23) | 0.87(0.21) | **0.88(0.21)** |
| Recall | 0.95(0.11) | 0.96(0.06) | 0.94(0.09) | **0.96(0.10)** |
| F-1 | 0.90(0.20) | 0.91(0.20) | 0.90(0.19) | **0.92(0.19)** |
| Segmentation - Dice Coefficient | | | | |
| All Videos | 0.81(0.12) | 0.83(0.12) | 0.83(0.11) | **0.84(0.11)** |
| SCBP Videos | 0.78(0.10) | 0.80(0.10) | 0.79(0.09) | **0.81(0.09)** |
| SCBP Pos. Frames | 0.74(0.10) | 0.76(0.10) | 0.75(0.09) | **0.77(0.09)** |
| ***Interscalene BP (ISC)*** | | | | |
| Object Detection at 50% IOU | | | | |
| Precision | 0.57(0.36) | 0.59(0.31) | 0.59(0.35) | **0.60(0.36)** |
| Recall | **0.85(0.28)** | 0.87(0.23) | 0.79(0.36) | 0.81(0.34) |
| F-1 | 0.68(0.36) | 0.70(0.32) | 0.68(0.37) | **0.69(0.36)** |
| Segmentation - Dice Coefficient | | | | |
| All Videos | 0.93(0.15) | 0.94(0.16) | 0.93(0.15) | **0.94(0.15)** |
| ISC Videos | 0.64(0.19) | **0.69(0.19)** | 0.65(0.19) | 0.68(0.19) |
| ISC Pos. Frames | 0.59(0.20) | **0.64(0.19)** | 0.61(0.20) | 0.63(0.19) |

Data Reported as mean (±SD)

TABLE III. UNet++ Performance on Different Machines

| | **Sonosite M-Turbo™** | | **Butterfly IQ™** | |
|---|---|---|---|---|
| | ***Before FT*** | ***After FT*** | ***Before FT*** | ***After FT*** |
| Object Detection at 50% IOU | | | | |
| Precision | 0.80(0.24) | 0.95(0.05) | 0.23(0.26) | 0.60(0.25) |
| Recall | 0.90(0.09) | 0.98(0.04) | 0.42(0.40) | 0.96(0.06) |
| F-1 | 0.85(0.19) | 0.96(0.04) | 0.30(0.30) | 0.74(0.19) |
| Segmentation - Dice Coefficient | | | | |
| All Videos | 0.69(0.09) | 0.78(0.03) | 0.55(0.17) | 0.67(0.08) |
| Pos. Frames | 0.69((0.08) | 0.77(0.03) | 0.45(0.16) | 0.66(0.08) |

FT: Fine-Tuning
Data Reported as mean (±SD)

### B. Needle Trajectory

After fine-tuning on 937 frames from 5 random videos, we tested the SCBP and needle on 124 frames in the sixth video. The needle trajectory was evaluated by comparing the ground truth and prediction for trajectory angle, trajectory's distance from center of the image, and needle tip coordinates.

The needle trajectory's average angle error was $1^o$, average error in trajectory's distance from center of the image was 10 pixels (2 mm) and the average error in needle tip was 15 pixels (3 mm) which is within acceptable range of 5 mm as per experienced anesthesiologists. The corresponding standard deviations were $1^o$, 7 pixels (1.4 mm) and 10 pixels (2 mm) respectively.

The average run-rate was measured to be around 24 ms using the NVIDIA RTX 3090 GPU and AMD Ryzen 1900X CPU.

## V. Discussion and Conclusion

In this study, we examined the computer vision related aspects of developing an autonomous robot capable of UGRA i.e. ultrasound image interpretation, definition and localization of the needle target, localization of needle and its tip and guidance of the needle towards the needle target.

First, we collected 227 ultrasound videos data from different patients. and created a large dataset comprising of 41,000 frames with manually vetted annotations of brachial plexus. We shortlisted four off-the-shelf segmentation models based on established usage in medical image segmentation literature, real-time processing capability and low memory usage. We tested the performance of each of these models with 5-fold cross-validation. For testing the generalizability of the model, we tested it on data from two separate US scanners - a traditional machine with similar image quality to the original machine in the main dataset, and an inexpensive handheld portable scanner with a compromise of a lower image quality. We achieved an acceptable dice score of brachial plexus segmentation with the similar quality machine but lower score with the lower image quality machine. The lower quality machine dice score was not unexpected. However, the significant improvement in dice score with slight fine-tuning suggests that the scores can be improved with more data.

Next we focus on the needle guidance towards needle targets. We annotated needles in six SCBP videos comprising of 1,061 frames. First, a natural images pre-trained UNet++ was fine-tuned on a large dataset of 196 US videos wherein the network learned the US domain and to segment SCBP and ISC nerves as two output channels. Next, a second fine-tuning on five needle videos was done to adapt the second output channel of ISC to predict needle mask. After training on five random videos we tested the needle in the sixth video.

We define automated anesthesia needle targets derived from CNN nerve segmentation predictions by fitting an ellipse to the nerve contours. Radon transform was applied on the network's needle mask prediction to obtain a sinogram. Using the angle derived from the sinogram and the lowest point in the needle mask, a line is drawn from the needle tip to show the needle trajectory. Finally, the intersection of extrapolated trajectory with the needle target guides the needle navigation for drug delivery.

We have publicly released both the datasets discussed above for further research by the research community. To the best of our knowledge, these are the largest publicly released ultrasound video datasets created from in-vivo human subjects containing brachial plexus annotations and needle annotations.

Algorithms utilizing spatio-temporal information should be investigated to exploit the temporal information. This can help in cases when needles get blurred or when they go out of field-of-view. For generalizability on new US machine, testing on other machines is required and some machines may require fine-tuning using their own data. The model may not be suitable for patients with severe obesity, whose nerves are situated deeper beneath the skin, as our dataset lacked patients with BMI>31. The study can be extended to other nerve blocks. This work focused on interpreting ultrasound images, locating the needle target, and guiding the needle to the target autonomously. While the accuracy attained is deemed satisfactory by anesthesiologists, further research is necessary to evaluate its acceptability for visual servoing and to develop an autonomous robotic system that can deliver anesthetics effectively in real world environments while minimizing unintended trauma to important contiguous organs.